\documentclass[10pt,twocolumn,letterpaper]{article}

\usepackage{cvpr}              % To produce the CAMERA-READY version
\usepackage{graphicx}
\usepackage{amsmath}
\usepackage{amssymb}
\usepackage{booktabs}
\usepackage{enumitem}
\usepackage{color}
\usepackage[pagebackref,breaklinks,colorlinks]{hyperref}

\usepackage[capitalize]{cleveref}
\crefname{section}{Sec.}{Secs.}
\Crefname{section}{Section}{Sections}
\Crefname{table}{Table}{Tables}
\crefname{table}{Tab.}{Tabs.}

\def\cvprPaperID{4814} % *** Enter the CVPR Paper ID here
\def\confName{CVPR}
\def\confYear{2022}

\begin{document}

% \title{R(Det)$^2$: Randomized Routing of Decision Trees for Object Detection}
\title{R(Det)$^2$: Randomized Decision Routing for Object Detection}

\author{Ya-Li Li \qquad \qquad \qquad \qquad Shengjin Wang\thanks{Corresponding author}\\
 Department of Electronic Engineering, Tsinghua University and BNRist, Beijing, China\\
 {\tt\small liyali13,wgsgj@tsinghua.edu.cn}
% 	\and
% Shengjin Wang\\
}
\maketitle

%%%%%%%%% ABSTRACT
\begin{abstract}
In the paradigm of object detection, the decision head is an important part, which affects detection performance significantly. Yet how to design a high-performance decision head remains to be an open issue. In this paper, we propose a novel approach to combine decision trees and deep neural networks in an end-to-end learning manner for object detection. First, we disentangle the decision choices and prediction values by plugging soft decision trees into neural networks. To facilitate effective learning, we propose randomized decision routing with node selective and associative losses, which can boost the feature representative learning and network decision simultaneously. Second, we develop the decision head for object detection with narrow branches to generate the routing probabilities and masks, for the purpose of obtaining divergent decisions from different nodes. We name this approach as the randomized decision routing for object detection, abbreviated as R(Det)$^2$. Experiments on MS-COCO dataset demonstrate that R(Det)$^2$ is effective to improve the detection performance. Equipped with existing detectors, it achieves $1.4\sim 3.6$\% AP improvement. % Code will be released soon.
\end{abstract}

%%%%%%%%% BODY TEXT
\section{Introduction}
\label{sec:intro}

\begin{figure}[htbp]
	\centering
	\includegraphics[width=3.2in]{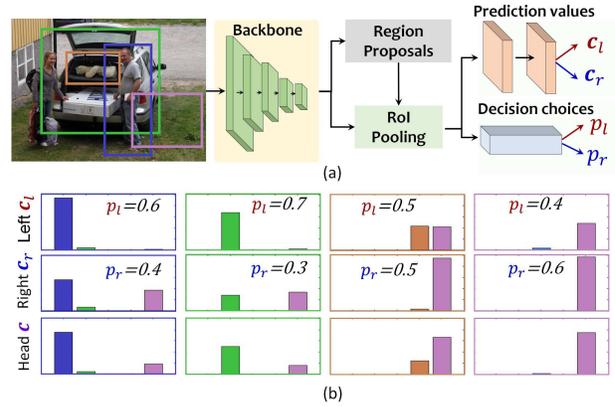}
	\caption{\textbf{Overview of the proposed approach.} (a) Inspired by decision trees, we disentangle the decision choices and predictive values by introducing tree structure for decision head in object detection. With multi-node prediction, we can explore more diverse cues. (b) We use the soft probability to denote decision choices for different routes of nodes. The overall decision is the weighted sum of prediction values from different nodes. Specially, we propose randomized decision routing to learn divergent decisions from different nodes for overall performance improvement. {\label{fig_outline}}}
\end{figure}

Object detection, which aims to recognize and localize the objects of interest in images, is a fundamental yet challenging task in computer vision. It is important for various applications, such as video surveillance, autonomous driving, and robotics vision. Due to its practical importance, object detection has attracted significant attention in the community. In recent decades, deep neural networks (DNNs) have brought significant progress into object detection. Typically, existing deep learning-based detection methods include one-stage detectors~\cite{Redmon_yolo16,Liu_ssd14,Lin_retina17}, two-stage detectors~\cite{Girshick_rcnn14,Ren_fasterrcnn15,Dai_rfcn16, Cai_cascadeRCNN18, Pang_libra19}, end-to-end detectors~\cite{Carion_detr20,Zhu_deformdetr21,Sun_tsprcnn21}.

Generally, current deep architectures constructed for object detection involve two components. One is the backbone for feature extraction, which can be pre-trained with large-scale visual recognition datasets such as ImageNet~\cite{Russakovsky_imagenet14}. The other is the decision head, which produces the predictions for computing losses or inferring detection boxes. Collaborated with region sampling, object detection can be converted into a multitask learning issue, where the decision tasks include classification and bounding box (\textit{bbox}) regression. For existing detection networks, the decision head is simply constructed by sequentially connecting several convolution or fully-connected layers. For one-stage detectors, the decision head is commonly constructed by stacking several convolutional layers. The decision head for region proposal in two-stage detectors is similar. For two-stage detectors, the region-wise decision in R-CNN stage is typically implemented with 2 fully-connected layers. Since the decision head is quite important for high-performance detectors, there are recently devoted researches~\cite{Wu_doublehead20,Song_tsd20,Dai_dyhead21,Feng_TOOD21}. However, most of these works focus on task disentanglement and task-aware learning, leaving the universal decision mechanism far from exploitation.

Considering that the features from DNNs show great potential for high-level vision tasks, the simple design of widely-adopted single-node decision might impede the performance of object detection. A natural question arises: \textit{is single-node prediction good enough for feature exploration in object detection?} To answer this, we focus on novel decision mechanism and propose an approach to introduce soft decision trees into object detection. As in Figure~\ref{fig_outline}, we integrate soft decision trees to disentangle the routing choices and prediction values. To jointly learn the soft decision trees and neural networks in an end-to-end manner, we propose the randomized decision routing with the combination of so-called \textit{selective loss} and \textit{associate loss}. Experiments validate the effectiveness of the proposed approach and address the necessity of introducing multi-node predictions. Since our work is mainly on \textbf{R}andomized \textbf{De}cision rou\textbf{t}ing for object \textbf{Det}ection, we name it as R(Det)$^2$. From the perspective of machine learning, our R(Det)$^2$ is an attempt to bridge the neural networks and decision trees – two mainstream algorithms, which would bring insights into future research.

The contributions of this paper are three-fold. 
\begin{itemize}[topsep=1pt, parsep=0pt, itemsep=0.5pt, partopsep=0pt]
	\item We propose to disentangle the route choices and prediction values for multi-node decision in object detection. In particular, we propose randomized decision routing for the end-to-end joint learning of the tree-based decision head. 
	\item We construct a novel decision head for object detection, which introduces
	routing probabilities and masks to generate divergent decisions from multiple nodes for the overall decision boosting.
	\item Extensive experiments validate the effectiveness of our proposed R(Det)$^2$. In particular, R(Det)$^2$ achieves over 3.6\% of $AP$ improvement when equipped with Faster R-CNN. It improves the detection accuracy of large objects by a large margin as well.
	
\end{itemize}

\section{Related work} \label{sec:rel}
\textbf{One-stage detectors.} Overfeat~\cite{Sermanet_overfeat13} predicts the decision values for classification and localization directly with convolutional feature maps. YOLO~\cite{Redmon_yolo16,Redmon_yolov3_18} regresses the object bounds and category probabilities directly based on image gridding. SSD~\cite{Liu_ssd14} improves the one-stage detection with various scales of multilayer features. Retina Net~\cite{Lin_retina17} proposes the focal loss to tackle the foreground-background imbalance issue. Besides, keypoints-based one-stage detectors~\cite{Law_corner18, Zhou_objects19, Duan_centernet19, Chen_reppoints20} have been extensively studied. CornerNet~\cite{Law_corner18} generates the heatmaps of top-left and bottom-right corners for detection. CenterNet~\cite{Duan_centernet19} uses a triplet of keypoints for representation with additional center points. Moreover, FCOS~\cite{Tian_FCOS19} and ATSS~\cite{Zhang_atss20} introduce centerness branch for anchor-free detection. Other methods delve into sample assignment strategies \cite{Zhang_atss20, Zhu_fsaf19, Cao_pisa20, Kong_foveabox20, Ge_OTA21, Ma_IQDet21}.

\textbf{Two-stage detectors.} R-CNN~\cite{Girshick_rcnn14}, Fast R-CNN~\cite{Girshick_fastrcnn15}, Faster R-CNN~\cite{Ren_fasterrcnn15} predict object scores and bounds with pooled features of proposed regions. R-FCN~\cite{Dai_rfcn16} introduces position-sensitive score maps to share the per-ROI feature computation. Denet~\cite{Tychsen_denet17} predicts and searches sparse corner distribution for object bounding. CCNet~\cite{Ouyang_ccnet17} connects chained classifiers from multiple stages to reject background regions. Cascade R-CNN~\cite{Cai_cascadeRCNN18} uses sequential R-CNN stages to progressively refine the detected boxes. Libra R-CNN~\cite{Pang_libra19} mainly tackles the imbalance training. Grid R-CNN~\cite{Lu_grid19} introduces pixel-level grid points for predicting the object locations. TSD~\cite{Song_tsd20} decouples the predictions for classification and box bounding with the task-aware disentangled proposals and task-specific features. Dynamic R-CNN~\cite{Zhang_dynamicrcnn20} adjusts the label-assigning IoU thresholds and regression hyper-parameters to improve the detection quality. Sparse R-CNN~\cite{Sun_sparsercnn21} learns a fixed set of sparse candidates for region proposal. 

\textbf{End-to-end detectors.} DETR~\cite{Carion_detr20} models object detection as a set prediction issue and solve it with transformer encoder-decoder architecture. It inspires the researches on transformer-based detection frameworks~\cite{Dai_updetr21,Zhu_deformdetr21,Liu_wbdetr21,Dai_dydetr21, Sun_tsprcnn21}. Deformable DETR~\cite{Zhu_deformdetr21} proposes the sparse sampling for key elements. TSP~\cite{Sun_tsprcnn21} integrates FCOS and R-CNN head into set prediction issue for faster convergence.

\textbf{Decision mechanism.} The decision head in object detection frameworks usually involves multiple computational layers (\textit{i.e.}, convolution layers, fully-connected layers and transformer modules). Typically, for one-stage detectors with dense priors~\cite{Redmon_yolo16,Liu_ssd14,Lin_retina17,Duan_centernet19,Tian_FCOS19}, stacked convolutions are used to obtain features with larger receptive fields, with separate convolution for classification, localization and other prediction tasks. For the decision in R-CNN stages~\cite{Ren_fasterrcnn15,Cai_cascadeRCNN18,Pang_libra19,Zhang_dynamicrcnn20,Lu_grid19}, stacked fully-connected layers are common. Double-head R-CNN~\cite{Wu_doublehead20} uses fully-connected layers for position-insensitive classification and fully-convolutional layers for position-sensitive localization. Dynamic head~\cite{Dai_dyhead21} unifies the scale-, spatial- and task-aware self-attention modules for multitask decisions.

% In this paper, we delve into the decision mechanism for object detection. We propose to disentangle the decision choice and values with the soft decision trees. Moreover, the randomized decision routing is proposed to enhance the feature representation and diversify the decision from different routes. Our propose R(Det)$^2$ can be easily integrated into existing detectors. Experiments show that R(Det)$^2$ achieves significant improvement compared to existing decision heads.
 
\section{Randomized decision trees}\label{sec:rdt}
\subsection{Soft decision trees}

To disentangle the decision choices and prediction values, we first construct soft decision trees~\cite{Hinton_soft17} for multiclass classification and \textit{bbox} regression in object detection. We use the soft routing probability ranging from 0 to 1 to represent the decision choice and facilitate network optimization.

\textbf{Soft decision tree for classification.} For multiclass classification, the soft decision tree is formulated as:
\begin{equation}
\mathbf{c}=\sum_{j\in \textit{Nodes}} p_{j} \mathbf{c}_{j}, \sum_{j \in \textit{Nodes}}p_{j}=1
\label{eq3_1}
\end{equation}
where $\mathbf{c}$ is the output of the whole classification tree and $\mathbf{c}_{j}$ is the prediction value from each node. $p_{j}$ is the routing probability for decision choice. It indicates the probability of choosing $j$-th classification node. For all the nodes, $\sum_{j \in \textit{Nodes}} p_{j}=1$. Eqn.~\ref{eq3_1} shows that $\mathbf{c}$ is the weighted sum of the classification scores from all the nodes. Different from traditional decision tree, $p_j$ is "soft" ranging from 0 to 1. $p_j$ can be obtained in networks by a scalar score with activations such as \textit{Softmax}, \textit{Sigmoid}. 

\textbf{Soft decision tree for regression.} For \textit{bbox} regression, we formulate the soft decision tree in a similar way as:
\begin{equation}
\mathbf{b}=\sum_{j\in \textit{Nodes}} q_{j} \mathbf{b}_{j}, \sum_{j \in \textit{Nodes}}q_{j}=1
\label{eq3_2}
\end{equation}
where $\mathbf{b}_j$ is the regression value output from each node $j$. $q_{j}$ is the routing probability for the $j$-th regression node. $\mathbf{b}$ is the output of the tree regressor. Similar to soft classification tree, the routing probability $q_{j} \in [0,1]$ is “soft”.

Noting that the routing probabilities $p_{j}$, $q_{j}$ denote decision choices, which indicates the probability of routing the $j$-th node. It can be viewed as decision confidence in test phase. $\mathbf{c}_{j}$ and $\mathbf{b}_{j}$ are the prediction values for classification and regression tasks attached with the $j$-th node. Both the decision choices and prediction values can be easily obtained with neural layers. With soft decision trees, multiple discriminative and divergent decisions can be obtained with features from different aspects. To facilitate the discussion, we restrict the soft decision tree as binary and $j \in \{l, r\}$.

\subsection{Randomized Decision Routing}

To learn soft decision trees in neural networks, we propose randomized decision routing. The motivation is two-fold. First, in order to obtain a high-performance decision head with tree structure, we need to avoid the high relevance of multiple predictions from different nodes. It means that we should differentiate the training to reduce the decision relevance of different nodes. Second, we also need to guarantee the decision performance of the whole tree. In a word, we need to achieve high-performance tree decision with low-relevant node decisions. To realize this, we propose the \textit{selective loss} to supervise the per-node learning and \textit{associative loss} to guide the whole-tree optimization. We then unify the \textit{selective} and \textit{associative loss} into a general training framework. Since we involve random factors to model the probability of routing different nodes, we name this training strategy as randomized decision routing.

\begin{figure}[htbp]
	\centering
	\includegraphics[width=3.0in]{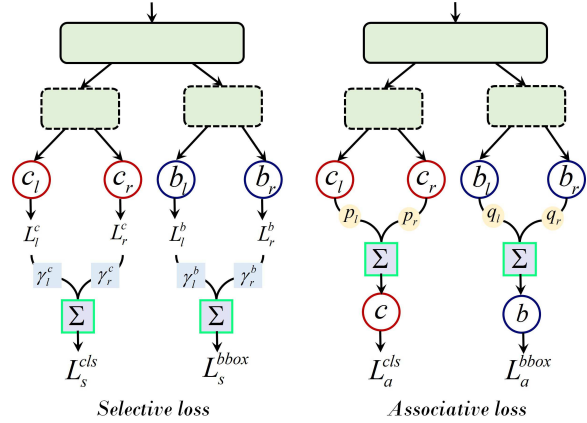}
	\caption{\textbf{Illustration on training deep networks with decision tree head.} We propose randomized decision routing which includes selective and associative losses. The selective loss identifies the dominant decisive prediction and weights the node loss accordingly in a randomized way. The associate loss learns the routing probability by measuring the difference between the fused output and the ground truth. {\label{fig_rr}}}
\end{figure}

% For instance, we choose the node with prediction relatively closer to the ground truth.
To achieve node decisions with low relevance, we first perform node selection to identify the node with higher optimization priority. We then attach the selected node with a higher routing probability. Oppositely, a lower routing probability is attached with the remaining node. Divergent routing probabilities lead to different learning rates for different nodes. Therefore, to diversify the decision of different nodes, we construct the \textit{selective loss} by setting different randomized weights for different node losses. As illustrated in Figure~\ref{fig_rr}-left, the \textit{selective losses} for classification and \textit{bbox} regression are denoted as:
\begin{equation}\label{eq3_3}
\begin{aligned} 
L_s^{cls}(\mathbf{c}_l, &\mathbf{c}_r,y)= \gamma_l^c L_l^c + \gamma_r^c L_r^c \\
&=\gamma_l^c L^{cls}(\mathbf{c}_l, y) + \gamma_r^c L^{cls} (\mathbf{c}_r, y) \end{aligned}
\end{equation}
\vspace{-1.5em}
\begin{equation}
\begin{aligned}\label{eq3_4}
L_s^{bbox}(\mathbf{b}_l, &\mathbf{b}_r,B)=\gamma_l^b L_l^b + \gamma_r^b L_r^b \\
&=\gamma_l^b L^{bbox}(\mathbf{b}_l, B) + \gamma_r^b L^{bbox} (\mathbf{b}_r, B)
\end{aligned}
\end{equation}
where $y$ is the ground truth label and $B$ is the ground truth for \textit{bbox} regression. $\gamma_l^c, \gamma_r^c$ are the weights indicating the probability for selective routing of classification tree. $\gamma_l^b, \gamma_r^b$ are the weights indicating the probability for selective decision routing of \textit{bbox} regression tree.

\begin{figure*}[htbp]
	\centering
	\includegraphics[width=1.0\linewidth]{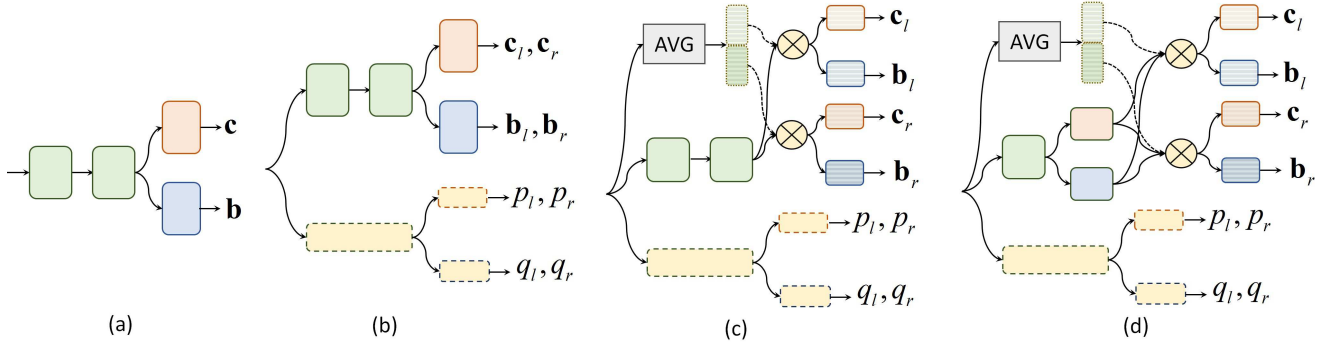}
	\caption{\textbf{Decision head for object detection.} (a) shows the common decision head. (b) shows \textit{R(Det)$^2$-B} which disentangles the decision choice and values by soft decision trees. (c) shows \textit{R(Det)$^2$-M} which leverages the routing masks to produce the divergent input features for decision. (d) shows \textit{R(Det)$^2$-T} which unifies task disentanglement into R(Det)$^2$-based decision head.  {\label{fig_dthead}}}
\end{figure*}

We leverage random weights to differentiate the node learning. For classification, we set $\gamma^c_l$, $\gamma^c_r$ based on the comparison of $L_l^c, L_r^c$. We set the nodes with lower loss values with higher random weights. For \textit{bbox} regression, we set the weights $\gamma^b_l$, $\gamma_r^b$ according to the relative comparison of $q_l, q_r$. For instance, if $q_l<q_r$, we restrict $\gamma_l^b<\gamma_r^b$. It is consistent with the intuition that we learn the selective node with higher priority in a fast way, meanwhile learning the remaining one in a slow way. Empirically, we sample the lower weight from $ U(0.1,0.3)$ and the higher weight from $U(0.9,1.1)$. This slow-fast randomized manner would benefit the learning of the whole decision head.

Besides of differentiating node decisions, we also need to ensure the performance of the whole decision tree. That is, the predictive decision output from the whole tree should be good. To achieve this, we formulate \textit{associative loss} based on the fused prediction $\mathbf{c}$, $\mathbf{b}$. The \textit{associative loss} can be the same as the original classification or \textit{bbox} regression loss in form, with the fused prediction as the input. As illustrated in Figure~\ref{fig_rr}-right, the \textit{associative loss} for classification and \textit{bbox} regression is formulated as:
\begin{equation}
L_a^{cls}\left(\mathbf{c},y\right)=L^{cls}\left(p_l \mathbf{c}_l+p_r \mathbf{c}_r,y\right)
\label{eq3_5}
\end{equation}
\vspace{-1.5em}
\begin{equation}
L_a^{bbox}\left(\mathbf{b},B\right)= L^{bbox}\left(q_l \mathbf{b}_l+q_r \mathbf{b}_r,B\right)
\label{eq3_6}
\end{equation}
The routing probabilities and prediction values are simultaneously optimized with the \textit{associative loss}. Specially, the routing probability which indicates the decision choice is only supervised by this \textit{associative loss}, resulting in appropriate routing in inference.

The whole loss is formulated as follows:
\begin{equation}
L_{all}=\lambda \left(L_s^{cls}+L_s^{bbox}\right)+(1-\lambda)\left(L_a^{cls}+L_a^{bbox}\right)
\label{eq3_7}
\end{equation}
where $\lambda \in [0,1]$ is the coefficient to balance between \textit{selective loss} and \textit{associative loss}. It is noteworthy that the $L^{cls}$, $L^{bbox}$ for computing the \textit{selective} and \textit{associative loss} can be commonly-used loss functions for classification (\textit{e.g.,} cross-entropy loss, Focal loss~\cite{Lin_retina17}) and \textit{bbox} regression (\textit{e.g.}, Smooth-L$_1$ loss, IoU loss~\cite{Yu_unitbox16,Tychsen_biou18,Rezatofighi_giou19,Zheng_diou20}). With soft decision trees, we can generate multiple decisions with different visual cues. Moreover, the divergent learning helps enhance feature representations and suppress over-optimization, further promote object detection.

\section{Decision head for detection}\label{sec:dthead}
We construct the head with decision trees for object detection. The common-used head of R-CNN detectors~\cite{Ren_fasterrcnn15, He_maskrcnn17,Cai_cascadeRCNN18, Li_lightrcnn17} is single-prediction type, as in Figure~\ref{fig_dthead}(a). Typically, two fully-connected (\textit{fc}) layers are sequentially connected with region-pooled features, with one additional \textit{fc} layer for classification and \textit{bbox} regression, respectively. In order to obtain decision values for multiple nodes, we first generate predictions $\mathbf{c}_l, \mathbf{c}_r$ and $\mathbf{b}_l, \mathbf{b}_r$ with the features output from the same structure as the common head. We further add another narrow branch with 1$\sim$2 \textit{fc} layers to produce the routing probabilities $p_l, p_r$ and $q_l, q_r$, as illustrated in Figure~\ref{fig_dthead}(b). We record this as the \textbf{B}asic head for randomized decision routing, as \textit{R(Det)$^2$-B}. The routing choices and predictions are disentangled with this basic head structure. 

Moreover, we add the routing masks for features before prediction to increase the divergence of decisions from multiple nodes. The decision values $\mathbf{c}_l, \mathbf{c}_r$ and $\mathbf{b}_l, \mathbf{b}_r$ are generated with route-wise masked features. As in Figure~\ref{fig_dthead}(c), we average the batched region-wise features to obtain a single context-like vector. Another \textit{fc} layer with \textit{Sigmoid} is imposed on this vector to produce routing masks for different nodes. By multiplying the route-wise masks on the last features before decision, we further diversify the input for different nodes of decision. The dependence of node decisions can be further reduced. We record this as \textbf{M}asked head for randomized decision routing, as \textit{R(Det)$^2$-M}. 

Inspired by efforts on disentangling the classification and localization tasks for detection, we develop another \textit{R(Det)$^2$-T}. We separate the last feature computation before the multitask prediction and unify the task-aware feature learning into our framework, as in Figure~\ref{fig_dthead}(d). Since it is not the main focus of this work, we have not involved more complicated task-aware head designs~\cite{Wu_doublehead20,Song_tsd20,Zhang_dynamicrcnn20}. Yet it is noteworthy that the proposed R(Det)$^2$ can easily be plugged into these detectors for performance improvement.

\section{Experiments}\label{sec:exp}

\textbf{Datasets.} We evaluate our proposed approach on the large-scale benchmark \textit{MS COCO} 2017~\cite{Lin_coco14}. Following common practice, we train detectors on \textit{training} split with $\sim$115k images and evaluate them on \textit{val} split with 5k images. We also report the results and compare with the state-of-the-art on \textit{COCO} \textit{test-dev} split with 20k images. The standard mean average precision (AP) across different IoU thresholds is used as the evaluation metric.

\textbf{Training details.} We implement the proposed R(Det)$^2$ as the plug-in head and integrate it into existing detectors. Our implementation is based on the popular mmdetection~\cite{Chen_mmdetection19} platform. If not specially noted, the R(Det)$^2$ serves for the decision in R-CNN of two-stage detectors, as Faster R-CNN~\cite{Ren_fasterrcnn15}, Cascade R-CNN~\cite{Cai_cascadeRCNN18}. We train the models with ResNet-50/ResNet-101~\cite{He_resnet16} backbones with 8 Nvidia TitanX GPUs. The learning rate is set to 0.02 and the weight decay is 1e-4, with momentum 0.9. The models for ablation studies are trained with the standard 1$\times$ configuration. No data augmentation is used except for standard horizontal image flipping. We only conduct multiscale training augmentation for evaluation on COCO \textit{test-dev} to compare with the state-of-the-art.

\textbf{Inference details.} It is noteworthy that the randomized decision routing is only performed in training phase. In inference, we perform on the single image scale without specific noticing. Following standard practice, we evaluate the models with test time augmentation (TTA) as multiscale testing to compare with the state-of-the-art.

\begin{table}[tbp]
	\centering
	\begin{tabular}{p{5mm}p{1mm}p{1mm}p{1mm}p{5mm}p{6mm}p{6mm}p{6mm}p{6mm}p{6mm}}
		\toprule[0.02in]
		&\textit{B} &\textit{M} &\textit{T} &$AP$ &$AP_{50}$ &$AP_{75}$ &$AP_S$ & $AP_M$ & $AP_L$ \\ 
		\midrule	
		\textit{2fc} & & & &37.4 &58.1 &40.4 &21.2 &41.0 &48.1 \\ \midrule
		\textit{2fc} &\checkmark & & &38.8 &59.8 &41.8 &22.3 &42.3 &50.9 \\
		& &\checkmark & &39.1 &60.5 &42.3 &22.5 &43.1 &50.5 \\
		& & &\checkmark
		&38.9 &60.2 &42.1 &23.1	&42.1 &50.2 \\ \midrule
		\textit{4conv} &\checkmark & & &38.7 &59.0 &41.9 &22.4 &42.0 &50.4 \\
		\textit{1fc}& &\checkmark & &39.2 &59.7 &42.4 &22.8 &42.8 &51.5 \\
		& & &\checkmark
		&39.5 &59.8 &42.9 &22.7	&43.1 &51.7 \\ \midrule
		\textit{4conv}&\checkmark & & &39.3 &60.2	&42.7 &22.5 &42.8 &51.6 \\
		\textit{(res)}& &\checkmark & &40.1 &60.8	&43.3 &23.3	&43.5 &52.6 \\
		\textit{1fc}& & &\checkmark &40.4 &61.2	&44.1 &23.8	&43.7 &53.0 \\ 		
		\bottomrule[0.02in]
	\end{tabular}
	\caption{\textbf{Ablation study on different types with R(Det)$^2$.} The baseline is Faster R-CNN equipped with ResNet-50 backbone. \textit{B}, \textit{M} and \textit{T} represents \textit{R(Det)$^2$-B}, \textit{R(Det)$^2$-M} and \textit{R(Det)$^2$-T} for decision heads, respectively.}
	\label{tab:ablation}
\end{table}

\subsection{Ablation study}

\textbf{Effects of components}.
We first conduct the ablative experiment to evaluate the effects of different components for R(Det)$^2$ (Table.~\ref{tab:ablation}). We integrate the proposed decision head structure into the R-CNN stage and apply randomized decision routing for training. We first follow the common setting with 2$\times$1024 fully-connected layers (referred as \textit{2fc}) to generate region-wise features, with decision values for multiclass classification and \textit{bbox} regression predicted based on them. By converting \textit{2fc} to R(Det)$^2$-B, we increase the detection $AP$ to 38.8\%, yielding 1.4\% of improvement. By adding routing masks for region-wise features, R(Det)$^2$-M achieves 39.1\% detection $AP$, 1.7\% of improvement. It is reasonable since the mask multiplying would promote the decision differences between nodes, leading to the improvement of joint decision. We further replace \textit{2fc} with 4$\times$256 convolutional layers with 1 fully-connected layer (referred as \textit{4conv1fc}). The achieved $AP$ increases to 38.7\%, 39.2\% and 39.5\% with R(Det)$^2$-B, R(Det)$^2$-M, R(Det)$^2$-T, respectively. We further add residual connections between neighboring convolutions for feature enhancement, referred to as \textit{4conv(res)1fc}. By integrating \textit{4conv(res)1fc} with R(Det)$^2$-B, we achieve $AP$ of 39.3\% and $AP_{75}$ of 42.7\%. By integrating R(Det)$^2$-M, the achieved $AP$ is 40.1\% and $AP_{75}$ is 43.3\%. With task disentanglement as R(Det)$^2$-T, we achieve $AP$, $AP_{50}$, $AP_{75}$ of 40.4\%, 61.2\% and 44.1\%, respectively. Compared to the baseline, the $AP$, $AP_{50}$, $AP_{75}$ is increased by 3.0\%, 3.1\% and 3.7\%, respectively. In particular, the R(Det)$^2$ significantly improves the detection accuracy on large objects, leading to the $AP_L$ improvement by a large margin. Compared with the baseline, we achieve 4.9\% of $AP_L$ improvement ultimately. It verifies that the features contain much more information to be exploited, especially for larger objects with high-resolution visual cues. Our proposed R(Det)$^2$ which produces decisions with multiple nodes can focus on the evidence from diverse aspects, leading to significant performance improvement.
\begin{table}[tbp]
	\centering
	\begin{tabular}{p{7mm}p{7mm}p{6mm}p{6mm}p{6mm}p{6mm}p{6mm}p{6mm}}
		\toprule[0.02in]
		$L^{cls}$ &$L^{bbox}$ &$AP$ &$AP_{50}$ &$AP_{75}$ &$AP_S$ & $AP_M$ & $AP_L$ \\ 
		\midrule	
		Baseline & &37.4 &58.1 &40.4 &21.2 &41.0 &48.1 \\ \midrule
		% \multicolumn{2}{c}{Baseline} &37.4 &58.1 &40.4 &21.2 &41.0 &48.1 \\
		CE &\small{S-L1}	&40.4 &61.2 &44.1 &23.8	&43.7 &53.0 \\
		Focal&\small{S-L1}	&40.5 &61.2	&44.4 &24.2 &43.6 &52.6 \\
		CE&IoU &40.9 &61.2	&44.5 &23.9	&44.2 &53.7 \\
		Focal&IoU &41.0 &61.1	&44.5 &24.3	&44.3 &53.7 \\
		\bottomrule[0.02in]
	\end{tabular}
	\caption{\textbf{Comparison with different loss functions.} The baseline model is Faster R-CNN with ResNet-50 as the backbone. CE indicates the cross-entropy loss. Focal indicates the original focal loss~\cite{Lin_retina17}. S-L1 indicates the Smooth-L$_1$ loss. IoU indicates the loss computed by the negative-log of intersection-over-union~\cite{Yu_unitbox16}.}
	\label{tab:loss}
\end{table}

\textbf{Effectiveness with different loss functions}.
The proposed randomized decision routing can be combined with any existing classification and localization losses. We conduct experiments to evaluate the effectiveness of R(Det)$^2$ with different loss functions(Table~\ref{tab:loss}). When we apply the \textit{Softmax} cross-entropy loss for classification and Smooth-L$_1$ loss for \textit{bbox} regression, we achieve 40.4\% $AP$, 61.2\% $AP_{50}$, 44.1\% $AP_{75}$. Compared to baseline Faster R-CNN with the same losses, we increase the $AP$, $AP_{50}$, $AP_{75}$ by 3.0\%, 3.1\%, 3.7\%, respectively. The $AP$ is slightly higher with focal loss~\cite{Lin_retina17} applying for classification. The detection $AP$ is further increased with IoU loss~\cite{Yu_unitbox16} applied for \textit{bbox} regression. The detection $AP$ reaches 41.0\%. Compared with the baseline, the $AP$ is increased by 3.6\% and $AP_L$ is increased by 5.6\%. It indicates that the proposed R(Det)$^2$ performs well with different combinations of loss functions, which further demonstrates its effectiveness.

\begin{table}[tbp]
	\centering
	\begin{tabular}{p{15mm}p{6mm}p{6mm}p{6mm}p{6mm}p{6mm}p{6mm}}
		\toprule[0.02in]
		Backbone &$AP$ &$AP_{50}$ &$AP_{75}$ &$AP_S$ & $AP_M$ & $AP_L$ \\ 
		\midrule
		R50 &37.4 &58.1 &40.4 &21.2 &41.0 &48.1 \\
		+R(Det)$^2$ &\textbf{41.0} &\textbf{61.2} &\textbf{44.8} &\textbf{24.6} &\textbf{44.1}	&\textbf{53.7} \\
		&\textbf{{\color{blue}(+3.6)}} &\textbf{{\color{blue}(+3.1)}} &\textbf{{\color{blue}(+4.4)}} &\textbf{{\color{blue}(+3.4)}} &\textbf{{\color{blue}(+3.1)}}	&\textbf{{\color{blue}(+5.6)}} \\
		\midrule
		\small{R50-DCN} &41.3 &62.4 &45.0 &24.6 &44.9 &54.4 \\
		+R(Det)$^2$ &\textbf{44.2} &\textbf{64.5} &\textbf{48.3} &\textbf{26.6}	&\textbf{47.7} &\textbf{58.6} \\
		&\textbf{{\color{blue}(+2.9)}} &\textbf{{\color{blue}(+2.1)}} &\textbf{{\color{blue}(+3.3)}} &\textbf{{\color{blue}(+2.0)}}	&\textbf{{\color{blue}(+2.8)}} &\textbf{{\color{blue}(+4.2)}} \\  \midrule
		% ResNet-50 (dcn+dpool) &44.6	&64.9 &48.7	&27.3 &48.0 &59.1 \\
		R101 &39.4 &60.1 &43.1 &22.4 &43.7 &51.1 \\
		+R(Det)$^2$ &\textbf{42.5} &\textbf{62.8}	&\textbf{46.3} &\textbf{25.1}	&\textbf{46.4} &\textbf{55.7} \\
		&\textbf{{\color{blue}(+3.1)}} &\textbf{{\color{blue}(+2.7)}}	&\textbf{{\color{blue}(+3.2)}} &\textbf{{\color{blue}(+2.7)}}	&\textbf{{\color{blue}(+3.7)}} &\textbf{{\color{blue}(+4.8)}} \\ \midrule
		\small{R101-DCN} &42.7 &63.7	&46.8 &24.9 &46.7 &56.8 \\
		+R(Det)$^2$ &\textbf{45.0} &\textbf{65.4} &\textbf{49.2} &\textbf{27.2} &\textbf{48.8} &\textbf{59.6} \\
		&\textbf{{\color{blue}(+2.3)}} &\textbf{{\color{blue}(+1.7)}} &\textbf{{\color{blue}(+2.4)}} &\textbf{\color{blue}{(+2.3)}} &\textbf{{\color{blue}(+2.1)}} &\textbf{{\color{blue}(+2.8)}} \\ 
	\bottomrule[0.02in]
	\end{tabular}
	\caption{\textbf{Comparison with different backbone networks.} R-50 and R-101 indicates ResNet-50 and ResNet-101, respectively. R(Det)$^2$ is plugged in Faster R-CNN with various backbones and achieves consistent performance gains.}
	\label{tab:backbone}
\end{table}

\textbf{Effectiveness on different backbone networks}.
With Faster R-CNN as the baseline detector, we conduct the ablative experiment to evaluate the effectiveness of R(Det)$^2$ on various backbones(Table~\ref{tab:backbone}). With ResNet-50 as the backbone, the achieved $AP$, $AP_{50}$ and $AP_{75}$ of R(Det)$^2$ is improved by 3.6\%, 3.0\%, and 4.1\%, respectively. With ResNet-50-DCN (ResNet-50 with deformable convolution) as the backbone, we achieve the detection $AP$ of 44.2\%, 2.9\% improvement. The performance gain of R(Det)$^2$ with ResNet-101 is also significant. By equipping with R(Det)$^2$, the detection $AP$ of ResNet-101 reaches 42.5\% and $AP_{75}$ reaches 46.3\%, 3.1\% and 3.2\% higher than the baseline. With ResNet-101-DCN as the backbone, the $AP$ reaches 45.0\% and $AP_{75}$ is 49.2\%. In particular, the detection accuracy over large objects is improved significantly. The $AP_L$ over the different backbones is increased by 5.6\%, 4.2\%, 4.8\% and 2.8\%, respectively. Experiments show that the proposed R(Det)$^2$ is effective among object detectors with various backbones.

\begin{table*}[tbp]
	\centering
	\begin{tabular}{p{36mm}p{16mm}p{18mm}p{18mm}p{18mm}p{18mm}p{18mm}}
		\toprule[0.02in]
		Detector &$AP$ &$AP_{50}$ &$AP_{75}$ &$AP_S$ & $AP_M$ & $AP_L$ \\ 
		\midrule 
		%Faster R-CNN~\cite{Ren_fasterrcnn15} &37.4 &58.1 &40.4 &21.2 &41.0 &48.1 \\
		% +R(Det)$^2$  &\textbf{40.4(+3.0)} &\textbf{61.2(+3.1)} &\textbf{44.1(+3.7)} &\textbf{23.8(+2.6)} &\textbf{43.7(+2.6)} &\textbf{53.0(+4.9)} \\ \midrule 
		% &41.0 &61.1 &44.5 &24.3 &44.3 &53.7 \\ \midrule
		Libra R-CNN~\cite{Pang_libra19} &38.3 &59.5 &41.9 &22.1 &42.0 &48.5 \\
		+R(Det)$^2$  &\textbf{41.4{\color{blue}(+3.1)}} &\textbf{61.4{\color{blue}(+1.9)}} &\textbf{45.5{\color{blue}(+3.6)}} &\textbf{24.7{\color{blue}(+2.5)}} &\textbf{45.0{\color{blue}(+3.0)}} &\textbf{53.7{\color{blue}(+5.2)}} \\  \midrule
		Cascade R-CNN~\cite{Cai_cascadeRCNN18} &40.3 &58.6 &44.0 &22.5	&43.8 &52.9 \\
		+R(Det)$^2$ &\textbf{42.5{\color{blue}(+2.2)}} &\textbf{61.0{\color{blue}(+2.4)}} &\textbf{45.8{\color{blue}(+1.8)}} &\textbf{24.6{\color{blue}(+2.1)}} &\textbf{45.5{\color{blue}(+1.7)}} &\textbf{57.0{\color{blue}(+4.1)}} \\ \midrule 
		% +R(Det)$^2$ &42.3(+2.0) &60.9(+2.3) &45.9(+1.9) &24.1(+1.6) &45.8(+2.0) &57.2(+4.3) \\
		Dynamic R-CNN~\cite{Zhang_dynamicrcnn20} &38.9 &57.6 &42.7 &22.1 &41.9 &51.7 \\
		+R(Det)$^2$ &\textbf{41.0{\color{blue}(+2.1)}} &\textbf{59.7{\color{blue}(+2.1)}} &\textbf{44.8{\color{blue}(+2.1)}} &\textbf{23.3{\color{blue}(+1.2)}} &\textbf{44.2{\color{blue}(+2.3)}} &\textbf{54.8{\color{blue}(+3.1)}} \\ \midrule[0.02in]
		DoubleHead R-CNN~\cite{Wu_doublehead20} &40.1 &59.4 &43.5 &22.9	&43.6 &52.9 \\
		+R(Det)$^2$ &\textbf{41.5(+1.4)} &\textbf{60.8(+1.4)} &\textbf{44.5(+1.0)} &\textbf{24.2(+1.3)} &\textbf{45.0(+1.4)} &\textbf{53.9(+1.0)} \\ 
		\bottomrule
	    RetinaNet~\cite{Lin_retina17} &36.5 &55.4 &39.1 &20.4	&40.3 &48.1 \\
		+R(Det)$^2$ &\textbf{38.3(+1.8)} &\textbf{57.4(+2.0)} &\textbf{40.8(+1.7)} &\textbf{22.6(+2.2)} &\textbf{42.0(+1.7)} &\textbf{50.5(+2.4)} \\ 
		\bottomrule[0.02in]
	\end{tabular}
	\caption{\textbf{Generalization with different detectors.} R(Det)$^2$ shows $AP$ improvement on various detectors. }
	\label{tab:detector}
\end{table*}

\textbf{Generalization on different detectors}. We plug R(Det)$^2$ into existing detectors to evaluate the generalization capability (Table~\ref{tab:detector}). Other than Faster R-CNN, we integrate R(Det)$^2$ with libra R-CNN~\cite{Pang_libra19}, dynamic R-CNN~\cite{Zhang_dynamicrcnn20}, cascade R-CNN~\cite{Cai_cascadeRCNN18}. The backbone is ResNet-50. Upon libra R-CNN, R(Det)$^2$ improves the detection $AP$ by 3.1\% and $AP_{75}$ by 3.6\%, yielding 41.4\% $AP$ and 45.5\% $AP_{75}$. On cascade R-CNN, the powerful detector with cascade structure, R(Det)$^2$ also shows consistent improvement. It improves the detection $AP$ by 2.2\% and $AP_{50}$ by 2.4\%, respectively. Since the dynamic R-CNN~\cite{Zhang_dynamicrcnn20} adaptively changes the hyperparameters of Smooth-L$_1$ loss for \textit{bbox} regression, we present the detection accuracy by randomized routing upon Smooth-L$_1$ loss, instead of IoU loss with better performance. By equipping R(Det)$^2$, the $AP$ and $AP_{75}$ is increased by 2.1\%. Besides, R(Det)$^2$ is quite effective to improve the detection performance of large objects. The $AP_L$ of libra R-CNN and cascade R-CNN is increased by a large margin with R(Det)$^2$, leading to 5.2\% and 4.1\% improvement, respectively. For DoubleHead R-CNN~\cite{Wu_doublehead20} and one-stage RetinaNet~\cite{Lin_retina17} with designed head, we fix the head for task-aware decision. Only randomized routing based training leads to 1.4\% of $AP$ improvement with DoubleHead R-CNN and 1.8\% of $AP$ improvement with RetinaNet~\cite{Lin_retina17}. The experiment validates that the proposed R(Det)$^2$ performs well on existing detectors.

\begin{figure}[htbp]
	\centering
	\includegraphics[width=1.0\linewidth,height=1.2in]{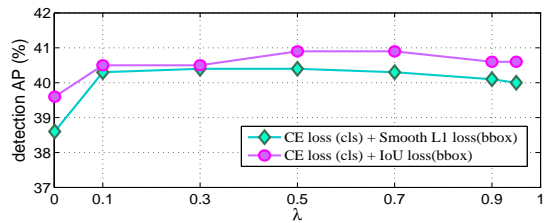}	
	\caption{\textbf{Effects on hyperparameter $\lambda$} to balance the \textit{selective loss} and \textit{associative loss} for decision routing.}
	\label{fig:param_lambda}
\end{figure} 

% \textbf{Hyperparameters}.
\textbf{Effects of hyperparameter $\lambda$}.
We leverage the hyperparameter $\lambda$ to balance the \textit{selective} and \textit{associative loss} in randomized decision routing. We further evaluate the effects of $\lambda$ with ResNet-50-based Faster R-CNN. The curves of detection $AP$ changing along with $\lambda$ are plotted in Figure~\ref{fig:param_lambda}. The detection accuracy is the highest when $\lambda=0.5$. That means we assign the weights for the \textit{selective} and \textit{associative loss} nearly equal. The detection $AP$ remains stable when $\lambda$ is between 0.1 to 0.9. If we further reduce $\lambda$ to 0.001 and reduce the impact of \textit{selective loss}, the detection $AP$ with Smooth-L$_1$ loss for \textit{bbox} regression decreases to 38.6\%, by 1.8\% points. It indicates that the \textit{selective loss} which aims to differentiate node decisions is essential for performance improvement. Since only associative loss guides the optimization of routing probabilities, increasing $\lambda$ to nearly 1 would lead to unstable models (the parameters to generate routing probabilities $p_l, p_r, q_l, q_r$ is nearly the same as random initialized ones), we restrict $\lambda \leq 0.95$. The detection $AP$ at $\lambda=0.95$ is decreased by 0.3$\sim$0.4\%.

\begin{table}[htbp]
	\centering
	\begin{tabular}{p{18mm}p{12mm}p{15mm}p{9mm}}
		\toprule[0.02in]
		\textit{Type} &\textit{\#FLOPs} &\textit{\#params} &$AP$(\%) \\ 
		\midrule	
		% \textit{4conv1fc} &321.72 &42.84 &37.6 \\ \midrule
		%\textit{R(Det)$^2$-B} &325.41 &46.53 &39.8 \\
		% \textit{R(Det)$^2$-M} &325.42 &53.10 &40.5 \\
		% \textit{R(Det)$^2$-T} &339.10 &73.19 &40.9 \\ 
		% \textit{R(Det)$^2$-Lite} &322.94 &45.70 &40.2 \\		
		\textit{4conv1fc} &129.0G &15.62M &37.6 \\ \midrule
		\textit{R(Det)$^2$-B} &132.6G &19.31M &39.8 \\
		\textit{R(Det)$^2$-M} &132.6G &25.88M &40.5 \\
		\textit{R(Det)$^2$-T} &146.3G &45.97M &40.9 \\ 
		\textit{R(Det)$^2$-Lite} &130.2G &18.48M &40.2 \\
		\bottomrule[0.02in]
	\end{tabular}
	\caption{Model complexity comparison of R(Det)$^2$ head.	}
	\label{tab:complexity}
\end{table}

\textbf{Model complexity and computational efficiency}. The model complexity of R(Det)$^2$ is mainly caused by the additional branches for routing probability, routing mask, and task-aware features. From Table~\ref{tab:complexity} we can see that the complexity is mainly caused by task-aware feature computation. Considering this, we develop \textit{R(Det)$^2$-Lite} with narrow computation for routing probabilities and masks, leading to 40.2\% $AP$ and nearly ignorable model complexity.

\begin{figure*}[htbp]
	\centering
	\includegraphics[width=0.96\linewidth]{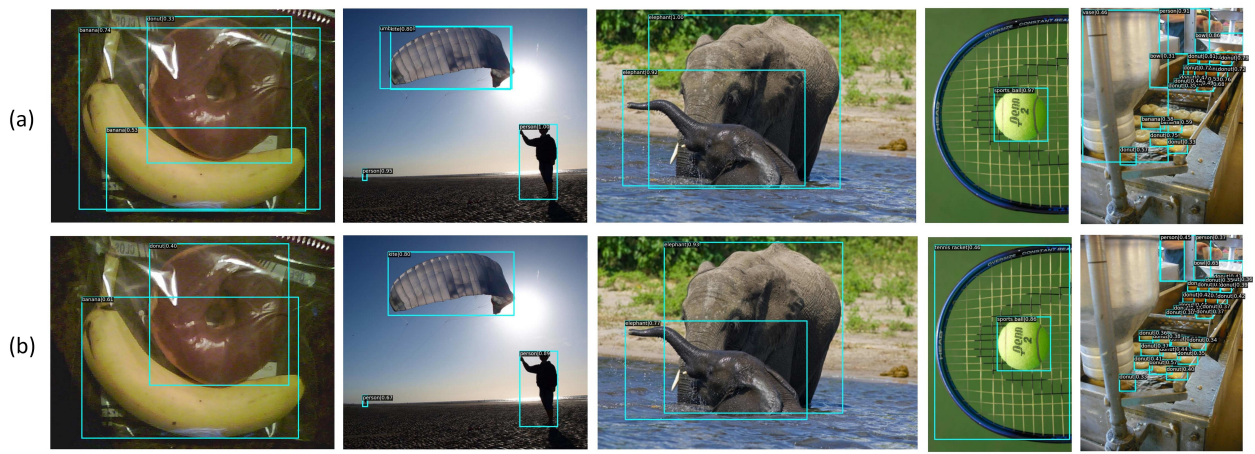}
	\caption{\textbf{Comparison of detection results for the baseline Faster R-CNN and R(Det)$^2$ equipped one.} The models are with ResNet-101 as the backbone and trained with \textit{COCO} 115k-\textit{train}. The example test images are from \textit{COCO} 5k-\textit{val}. The rectangles mark the detected bounding boxes with attached category labels and confidences. The detection results of baseline model are presented in (a) (39.3\% AP) and those of R(Det)$^2$ are presented in (b) (42.5\% AP). {\label{fig_vis101}}}
\end{figure*}

\textbf{Visualization.} We present the comparative visualization in Figure~\ref{fig_vis101}. The detected results by ResNet-101 based Faster R-CNN are shown in Figure~\ref{fig_vis101}(a) and those from the R(Det)$^2$ are shown in Figure~\ref{fig_vis101}(b). It can be seen that the proposed R(Det)$^2$ is effective to improve both the detection and localization performance. Specially, the R(Det)$^2$ is quite effective in reducing the repeated detections and avoiding over-confident ones.

\begin{table*}[htbp]
	\begin{center}
		\begin{small}
			\begin{tabular}{p{32mm}|p{35mm}|p{6mm}|p{5mm}|p{9mm}p{8mm}p{8mm}p{7mm}p{7mm}p{7mm}}
				\toprule[0.02in]
				Methods &Backbone &ME &TTA &$AP$ &$AP_{50}$ &$AP_{75}$ &$AP_S$ &$AP_M$ &$AP_L$ \\ \midrule
				%\textit{One-stage methods}& & &\multicolumn{6}{c}{} \\  \hline
				% YOLO-v3~\cite{Redmon_yolov3_18} &DarkNet-53 & & &33.0 &57.9 &34.4 &18.3 &35.4 &41.9 \\
				% SSD513~\cite{Liu_ssd14} &ResNet-101  & &31.2	&50.4 &33.3 &10.2 &34.5 &49.8 \\
				% DSSD513~\cite{Fu_DSSD17}	 &ResNet-101 & &33.2 &53.3 &35.2 &13.0 &35.4 &51.1 \\
				% Retina-Net~\cite{Lin_retina17} &ResNet-101 &39.1 &59.1	&42.3 &21.8	&42.7 &50.2 \\
				Retina-Net~\cite{Lin_retina17} &ResNeXt-101 &18e &  &40.8	&61.1 &44.1 &24.1 &44.2 &51.2 \\
				% GHM-Retina-Net~\cite{Li_ghm19} &ResNet-101 &39.9 &60.8 &42.5 &20.3 &43.6	&54.1 \\
				% GHM-Retina-Net~\cite{Li_ghm19} &ResNeXt-101 & & &41.6	&62.8 &44.2 &22.3	 &45.1 &55.3 \\
				% FSAF~\cite{Zhu_fsaf19} &ResNet-101  	&40.9 &61.5 &44.0 &24.0 &44.2	&51.3 \\
				% FSAF \textit{w ms}~\cite{Zhu_fsaf19} &ResNet-101  &42.8 &63.1 &46.5 &27.8 &45.5 &53.2 \\
				% FSAF~\cite{Zhu_fsaf19} &ResNeXt-101 &24e & &42.9 &63.8 &46.3 &26.6 &46.2 &52.7 \\
				FCOS~\cite{Tian_FCOS19} &ResNeXt-101 &24e & &43.2 &62.8 &46.6 &26.5 &46.2 &53.3 \\
				% Centernet~\cite{Duan_centernet19} &Hourglass-104 &100e & &44.9 &62.4 &48.1 &25.6 &47.4 &57.4 \\
				ATSS~\cite{Zhang_atss20} &ResNeXt-101-DCN &24e & &47.7 &66.5 &51.9 &29.7 &50.8 &59.4 \\ 
				OTA~\cite{Ge_OTA21} &ResNeXt-101-DCN &24e & &49.2 &67.6 &53.5 &30.0 &52.5 &62.3 \\ 
				IQDet~\cite{Ma_IQDet21} &ResNeXt-101-DCN &24e & &49.0 &67.5 &53.1 &30.0 &52.3 &62.0 \\  
				\midrule
				Faster R-CNN~\cite{Ren_fasterrcnn15}&ResNet-101 &12e &  &36.7 &54.8 &39.8 &19.2 &40.9 &51.6 \\
				% R-FCN~\cite{Dai_rfcn16} &ResNet-101  & & & 35.1 &52.9 &37.9 &16.4 &39.0 &50.7 \\
				% Deformable R-FCN~\cite{Dai_deform17}& Aligned-Inception-ResNet & & &37.5 &58.0 &40.8 &19.4 &40.1 &52.5 \\
				% Faster R-CNN \textit{w/FPN}~\cite{Lin_FPN17} &ResNet-101 &  &36.2 &59.1 &39.0	 &18.2 &39.0 &48.2 \\
				Libra R-CNN~\cite{Pang_libra19} &ResNeXt-101 &12e & &43.0 &64.0 &47.0 &25.3 &45.6 &54.6 \\
				Cascade R-CNN~\cite{Cai_cascadeRCNN18} &ResNet-101 &18e & &42.8 &62.1 &46.3 &23.7 &45.5 &55.2 \\
				% Dynamic R-CNN~\cite{Zhang_dynamicrcnn20} &ResNet-101-DCN &36e & &42.0 &60.7 &45.9 &22.7 &44.3 &54.3 \\  
				% TSD~\cite{Song_tsd20} &ResNet-101  & & &43.2 &64.0 &46.9 &24.0 &46.3 &55.8 \\ 
				TSP-RCNN~\cite{Sun_tsprcnn21} &ResNet-101-DCN &96e & &47.4 &66.7 &51.9 &29.0 &49.7 &59.1 \\
				Sparse R-CNN~\cite{Sun_sparsercnn21} &ResNeXt-101-DCN &36e  & &48.9 &68.3 &53.4 &29.9 &50.9 &62.4 \\ 
				Deformable DETR~\cite{Zhu_deformdetr21} &ResNeXt-101-DCN &50e & 	&50.1 &69.7 &54.6 &30.6 &52.8 &64.7 \\
				\midrule 
				Ours - R(Det)$^2$ &ResNeXt-101-DCN &12e & &\textbf{50.0} &\textbf{69.2} &\textbf{54.3} &\textbf{30.9} &\textbf{53.0} &\textbf{63.9} \\
				% Ours - R(Det)$^2$ &ResNeXt-101-DCN  &20e & &\textbf{51.3} &\textbf{70.4} &\textbf{55.8} &\textbf{32.0} &\textbf{54.3} &\textbf{65.4} \\
				% Ours - R(Det)$^2$ &ResNeXt-101-DCN  &36e & &\textbf{51.7} &\textbf{71.0} &\textbf{56.0} &\textbf{32.6} &\textbf{54.6} &\textbf{65.6} \\
				Ours - R(Det)$^2$ &Swin-L~\cite{Liu_swin21} &12e & &\textbf{55.1} &\textbf{74.1} &\textbf{60.4} &\textbf{36.0} &\textbf{58.6} &\textbf{70.0} \\ \midrule[0.01in] 
				% FSAF~\cite{Zhu_fsaf19} &ResNeXt-101-DCN &20e &\checkmark &44.6 &65.2 &48.6 &29.7 &47.1 &54.6 \\
				% FCOS~\cite{Tian_FCOS19} &ResNeXt-101-DCN &20e &\checkmark &44.7 &64.1 &48.4 &27.6 &47.5 &55.6 \\
				Centernet~\cite{Duan_centernet19} &Hourglass-104 &100e &\checkmark &47.0 &64.5 &50.7 &28.9 &49.9 &58.9 \\
				% SEPC~\cite{Wang_sepc20} &ResNeXt-101-DCN &24e &\checkmark &50.1 &69.8 &54.3 &31.3 &53.3 &63.7 \\
				ATSS~\cite{Zhang_atss20} &ResNeXt-101-DCN &24e &\checkmark &50.7 &68.9 &56.3 &33.2 &52.9 &62.4 \\
				% PAA~\cite{Kim_paa20} &ResNeXt-101-DCN &20e &\checkmark &51.3 &68.8 &56.6 &34.3 &53.5 &63.6 \\
				IQDet~\cite{Ma_IQDet21} &ResNeXt-101-DCN &24e &\checkmark &51.6 &68.7 &57.0 & 34.5 &53.6 &64.5 \\
				OTA~\cite{Ge_OTA21} &ResNeXt-101-DCN &24e &$\checkmark $ &51.5 &68.6 &57.1 &34.1 &53.7 &64.1 \\ 
				% PAA~\cite{Kim_paa20} &ResNeXt-152-DCN &24e &\checkmark &53.5 &71.6 & 59.1 &36.0 &56.3 &66.9 \\
				\midrule
				% SpineNet-190~\cite{Du_spinenet20} &SpineNet-190 & & &52.1 &71.8 &56.5 &35.4 &55.0 &63.6 \\
				% EfficientDet-D7~\cite{Tan_efficientdet20} &EfficientNet-B6 &  & &52.6 &71.6 &56.9 &- &- &- \\ \midrule
				% \textit{Two-stage methods}& & &\multicolumn{6}{c}{} \\  \hline
				Dynamic R-CNN~\cite{Zhang_dynamicrcnn20} &ResNet-101-DCN &36e &\checkmark &50.1 &68.3 &55.6 &32.8 &53.0 &61.2 \\
				% GCNet~\cite{Cao_gcnet19}&ResNeXt-101-DCN &  &48.4 &67.6 &52.7 &- &- &- \\
				% TSD~\cite{Song_tsd20} &ResNet-101-DCN &\checkmark &49.4 &69.6 &54.4 &32.7 &52.5 &61.0 \\
				TSD~\cite{Song_tsd20} &SENet154-DCN &36e &\checkmark  &51.2 &71.9 &56.0 &33.8 &54.8 &64.2\\ 
				Sparse R-CNN~\cite{Sun_sparsercnn21} &ResNeXt-101-DCN &36e  &\checkmark &51.5 &71.1 & 57.1 &34.2 &53.4 &64.1 \\ 
				% BorderDet~\cite{Qiu_Borderdet20} &ResNeXt-101-DCN &24e &\checkmark &50.3 &68.9 &55.2 &32.8 &52.8 &62.3 \\
				RepPoints v2~\cite{Chen_reppoints20} &ResNeXt-101-DCN &24e &\checkmark &52.1 &70.1 &57.5 &34.5 &54.6 &63.6 \\		
				Deformable DETR~\cite{Zhu_deformdetr21} &ResNeXt-101-DCN &50e &\checkmark &52.3 &71.9 &58.1 &34.4 &54.4 &65.6 \\
				RelationNet++~\cite{Chi_relation20} &ResNeXt-101-DCN &24e &\checkmark  &52.7 &70.4 &58.3 &35.8 &55.3 & 64.7 \\
				DyHead~\cite{Dai_dyhead21} &ResNeXt-101-DCN &24e &\checkmark  &54.0 &72.1 &59.3 &37.1 &57.2 &66.3 \\ \midrule
				
				% R(Det)$^2$（&ResNet-101 &  &44.4 &63.3 &48.5 &27.1 &47.9 &54.2 \\
				Ours - R(Det)$^2$ &ResNeXt-101-DCN &24e &\checkmark &\textbf{54.1} &\textbf{72.4} &\textbf{59.4} &\textbf{35.5} &\textbf{57.0} &\textbf{67.3} \\
				Ours - R(Det)$^2$ &Swin-L~\cite{Liu_swin21} &12e &\checkmark  &\textbf{57.4} &\textbf{76.1} &\textbf{63.0} &\textbf{39.4} &\textbf{60.5} &\textbf{71.5} \\ \bottomrule[0.02in]
			\end{tabular}
		\end{small}
		\caption{\textbf{Comparison of R(Det)$^2$ with the state-of-the-art object detection methods on \textit{COCO test-dev} dataset.} DCN indicates that using the deformable convolution to enhance the feature representations of backbone. TTA indicates test-time augmentation such as multi-scale testing and horizontal image flipping. ME indicates more epochs of training. } \label{table_coco_test}
	\end{center}
\end{table*}

\subsection{Comparison with the state-of-the-art}
We integrate the proposed R(Det)$^2$ into Cascade R-CNN to compare with the state-of-the-art methods on \textit{COCO} \textit{test-dev} dataset. The backbone is ResNeXt-101 (64$\times$4d)~\cite{Xie_resnext16} with deformable convolution and swin transformer~\cite{Liu_swin21}. The comparative study is presented in Table~\ref{table_coco_test}. We first compare the single-model single-scale model performance. With 12 epochs ($1\times$) of training, the R(Det)$^2$ achieves $AP$ of 50.0\%, outperforming Faster R-CNN~\cite{Ren_fasterrcnn15}, Libra R-CNN~\cite{Pang_libra19}, Cascade R-CNN~\cite{Cai_cascadeRCNN18} by a large margin. Compared with the recent Sparse R-CNN~\cite{Sun_sparsercnn21} with the same backbone, we achieve 1.1\% $AP$ improvement with 1/3 training iterations. It is also comparable with deformable DETR~\cite{Zhu_deformdetr21} with transformer architecture and much more epochs of training (50 epochs). The detection accuracy is further improved with more epochs of training and test-time augmentation as multi-scale testing and horizontal image flipping. With 24 epochs of training and TTA, the R(Det)$^2$ achieves $AP$ of 54.1\% and $AP_{50}$ of 72.4\%. Compared with DyHead with stacked self-attention modules~\cite{Dai_dyhead21}, the $AP_{50}$, $AP_L$ is improved by 0.3\% and 1.0\%, respectively. Besides, we adapt the backbone of ViT as swin transformer~\cite{Liu_swin21}. With 12 epochs of training, the achieved $AP$ of single-scale testing is 55.1\% and that of multi-scale testing is 57.4\%. It validates the R(Det)$^2$ performs well with various backbones and is effective for high-performance object detection.

\section{Conclusion}\label{sec:con}
The decision head is important for high-performance object detection. In this paper, we propose a novel approach as the randomized decision routing for object detection. First, we plug soft decision trees into neural networks. We further propose the randomized routing to produce accurate yet divergent decisions. By randomized routing for soft decision trees, we can obtain multi-node decisions with diverse feature exploration for object detection. Second, we develop the decision head for detection with a narrow branch to generate routing probabilities and a wide branch to produce routing masks. By reducing the relevance of node decisions, we develop a novel tree-like decision head for deep learning-based object detection. Experiments validate the performance of our proposed R(Det)$^2$.

\section*{Acknowledgement}
This work was supported by the state key development program in 14th Five-Year under Grant Nos.2021QY1702, 2021YFF0602103, 2021YFF0602102. 
We also thank for the research fund under Grant No. 2019GQG0001 from the Institute for Guo Qiang, Tsinghua University.

%%%%%%%%% REFERENCES
{\small
\bibliographystyle{ieee_fullname}
\bibliography{egbib}
}

\end{document}